\documentclass{article}
\usepackage{spconf,amsmath,graphicx}

\DeclareMathOperator*{\argmax}{arg\,max}
\usepackage{algorithm}
\usepackage[noend]{algpseudocode}
\usepackage{mathtools}
\usepackage{url}
\usepackage{makecell}
\usepackage{subcaption}
\usepackage{amssymb}
\usepackage{amsfonts}
\usepackage{dsfont}
\usepackage{amsthm}

\theoremstyle{definition}
\newtheorem{definition}{Definition}[section]

\title{L-RED: Efficient Post-Training Detection of Imperceptible Backdoor Attacks without Access to the Training Set}
\name{Zhen Xiang, David J. Miller and George Kesidis\thanks{Supported by an AFOSR DDDAS grant and Cisco URP gift.}
}
\address{Anomalee Inc. and School of EECS, Pennsylvania State University}
\begin{document}
%
\maketitle
\vspace{-0.05in}
\begin{abstract}
Backdoor attacks (BAs) are an emerging form of adversarial attack typically against deep neural network image classifiers. The attacker aims to have the classifier learn to classify to a target class when test images from one or more source classes contain a backdoor pattern, while maintaining high accuracy on all clean test images. Reverse-Engineering-based Defenses (REDs) against BAs do not require access to the training set but only to an independent clean dataset. Unfortunately, most existing REDs rely on an unrealistic assumption that all classes except the target class are source classes of the attack. REDs that do not rely on this assumption often require a large set of clean images and heavy computation. In this paper, we propose a Lagrangian-based RED (L-RED) that does not require knowledge of the number of source classes (or whether an attack is present). Our defense requires very few clean images to effectively detect BAs and is computationally efficient. Notably, we detect 56 out of 60 BAs using only two clean images per class in our experiments on CIFAR-10.
\end{abstract}
\vspace{-0.05in}
\vspace{-0.1in}
\section{Introduction}
\label{sec:intro}
\vspace{-0.1in}

Recently, a backdoor attack (BA) was proposed \cite{BadNet, Targeted, Trojan}, typically against deep neural network (DNN) image classifiers, where the attacker aims to: 1) induce test-time misclassification to a target class whenever a test image from one (or several) source class(es) is embedded with an attacker-specified backdoor pattern; 2) otherwise, induce correct test-time decision making. The attacker's goals can be achieved by poisoning the training set of the classifier with a relatively small number of ``backdoor training images'' -- images originally from one of the source classes, but with the attacker-specified backdoor pattern embedded and labeled to the target class. The attacker’s poisoning capability is facilitated particularly when practical training requires seeking data from public sources (which is easily attacked) \cite{Review}.

Defense against BAs can be deployed during the classifier's training, where the defender/learner has access to the training set and aims to cleanse it, while also using it for model building \cite{SS, AC, CI, Differential_Privacy}. However, classifiers threatened by BAs are usually part of downstream applications (e.g. widespread mobile apps) or legacy systems, for which the training set is {\em not accessible} to the defender/user. In this {\it ``post-training''} regime, the defender/user only has access to the trained classifier and whatever clean samples (a small set he/she can collect), and aims to detect whether the classifier has been backdoor attacked. Post-training defenses of \cite{SentiNet} and \cite{STRIP} infer if an input test image contains a backdoor pattern. In addition to the clean dataset for detection, both methods also require sample images containing the true backdoor pattern. The defense of \cite{Meta} trains a large number of backdoored and clean classifiers as available detection benchmarks, which requires substantial clean images and extensive computation.

A different group of post-training defenses, Reverse-Engineering-based Defenses (REDs), do not require access to the true backdoor pattern. Moreover, for these defenses, the number of clean images for detection are usually not sufficient to train even a shallow DNN. However, existing REDs either rely on an unrealistic assumption about the attack that the source classes include {\it all} classes except the target class \cite{NC, Tabor, SentiNet, DataLimited}, or require a significant number of clean images (and thus heavy computation) as compensation to relieve such assumption \cite{Post-ICASSP, MAMF}.

In this paper, we propose a computationally {\it efficient} Lagrangian-based RED (L-RED) to detect an {\it imperceptible} BA with {\it arbitrary} number of source classes, and infer the target class for any detected attack. Our defense can also be potentially extended with little modification to detect {\it perceptible} BAs. Compared with the state-of-the-art RED against imperceptible BAs with arbitrary number of source classes (\cite{Post-ICASSP}), our defense requires much fewer clean images for accurate BA detection (and thus also much lower computational complexity). Even with just two clean image per class, our defense detects 56 out of 60 attacks with 1 out of 10 false detections in our experiments on CIFAR-10.

\vspace{-0.1in}
\section{Background}
\label{sec:background}

\vspace{-0.1in}
\subsection{Imperceptible Backdoor Attack}
\label{subsec:backdoor}
\vspace{-0.05in}

A BA is typically specified by a target class with label $t^{\ast}\in{\mathcal C}$ ($|{\mathcal C}|=K$), a set of source classes ${\mathcal S}^{\ast}\subset{\mathcal C}$, where $t^{\ast}\notin{\mathcal S}^{\ast}$, and a backdoor pattern. Effective backdoor patterns in the literature are either {\it human-imperceptible} (\cite{Targeted, SS, Haoti, Post-ICASSP}) or {\it human-perceptible} (\cite{BadNet, NC, Tabor}). Here we focus on the imperceptible case, where the backdoor pattern is embedded into a clean image ${\bf x}\in{\mathcal X}$ by
\vspace{-0.1in}
\begin{equation}\label{eq:bd_imperceptible}
m({\bf x}; {\bf v}^{\ast}) = [{\bf x}+{\bf v}^{\ast}]_c
\end{equation}
with ${\bf v}^{\ast}$ an image-wide additive perturbation that is small in $||{\bf v}^{\ast}||_2$ for imperceptibility (and also possibly small in $||{\bf v}^{\ast}||_0$ for sparsity). Here, $[\cdot]_c$ is a domain-specific clipping function.

A BA is typically launched by poisoning the classifier's {\it training set} with a small set of images originally from the source classes ${\mathcal S}^{\ast}$, embedded with the backdoor pattern ${\bf v}^{\ast}$, and labeled to the target class $t^{\ast}$ \cite{BadNet, Targeted}. For a successful attack, the trained classifier $f(\cdot; \theta):{\mathcal X}\rightarrow{\mathcal C}$ should: 1) classify to the target class $t^{\ast}$ when any {\it test image} from any source class in ${\mathcal S}^{\ast}$ is embedded with ${\bf v}^{\ast}$; 2) maintain high classification accuracy on {\it clean} images during testing.

\vspace{-0.1in}
\subsection{Reverse-Engineering-Based Backdoor Defense (RED)}
\label{subsec:red}
\vspace{-0.05in}

REDs are post-training BA defenses {\it without access to the training set}, but with access to the {\it trained classifier} and an {\it independent, clean dataset} \cite{Post-ICASSP, NC}. REDs typically consist of a backdoor pattern reverse-engineering/estimation step and an anomaly detection step. For {\it imperceptible} BA detection, the key ideas of \cite{Post-ICASSP} are: 1) for any ``non-backdoor'' class pair $(s, t)\in{\mathcal C}\times{\mathcal C}$ where $t\neq t^{\ast}$ or $s\notin{\mathcal S}^{\ast}$, and a {\it sufficiently large} set ${\mathcal D}_s$ of clean images from class $s$, the {\it min-norm} perturbation required (via (\ref{eq:bd_imperceptible})) to induce most images in ${\mathcal D}_s$ to be (mis)classified to $t$ is {\it large}; 2) there exists a {\it small-norm} perturbation (ensured by the existence of $||{\bf v}^{\ast}||$) that induces most images in ${\mathcal D}_s$ to be (mis)classified to $t^{\ast}$ for any $s\in{\mathcal S}^{\ast}$. 
Thus, the estimation step in \cite{Post-ICASSP} searches for the minimum $l_2$ norm pattern inducing at least $\pi$-fraction of misclassifications on ${\mathcal D}_s$, for each $(s, t)\in{\mathcal C}\times{\mathcal C}$ ($s\neq t$) class pair:
\vspace{-0.05in}
\begin{equation}\label{eq:opt_raw_imperceptible}
\begin{aligned}
& \underset{\bf v}{\text{minimize}}
& & ||{\bf v}||_2\\
\vspace{-0.1in}
& \text{subject to}
& & \frac{1}{|\mathcal{D}_s|}\sum_{{\bf x}\in\mathcal{D}_s}{\mathds 1} (f(m({\bf x};{\bf v});\theta)=t) ~\geq~ \pi,
\end{aligned}
\vspace{-0.05in}
\end{equation}
where ${\mathds 1}(\cdot)$ is an indicator function, and $\pi\in(0, 1)$ is chosen to be large (e.g. $\pi=0.9$). Then ``backdoor'' class pairs can be revealed by an anomaly detector because the pattern obtained for any ``backdoor'' class pair should have an {\it abnormally small} norm compared with the pattern obtained for ``non-backdoor'' class pairs. Note that REDs can also target {\it perceptible} BAs that use a local patch replacement (e.g. glasses on a face) rather than an additive perturbation (\ref{eq:bd_imperceptible}) \cite{NC, MAMF}. But the pattern estimation problem formulation for these REDs is similar to (\ref{eq:opt_raw_imperceptible}) (see Appx. \ref{apdx:perceptible}). REDs against imperceptible BAs and perceptible BAs can be deployed in parallel, to provide a complete solution covering both cases.

{\bf Limitations of existing REDs:} Some REDs, e.g. \cite{NC, Tabor, DataLimited}, assume that {\it the source classes consist of all classes except the target class}, i.e. ${\mathcal S}^{\ast}\cup t^{\ast}={\mathcal C}$. Correspondingly, their pattern estimation is performed for each putative target class $t$ using the union $\cup_{s\neq t}{\mathcal D}_s$ (instead of for each class pair $(s, t)$, $s\neq t$). However, any mismatch between the assumed source classes used by the defender and the actual source classes used by the attacker will hurt detection accuracy (as will be shown experimentally). Moreover, it may be impractical for attacker to collect images from all classes other than $t^{\ast}$, for creating backdoor training images, especially when $K=|{\mathcal C}|$ is large.
In contrast, REDs like \cite{Post-ICASSP} perform pattern estimation (\ref{eq:opt_raw_imperceptible}) for all $K(K-1)$ class pairs to handle detection of BAs with {\it arbitrary} $|{\mathcal S}^{\ast}|$. However, this method requires $|{\mathcal D}_s|$ in (\ref{eq:opt_raw_imperceptible}) to be {\it sufficiently large} to have a significant outlier for ``backdoor'' class pairs (otherwise, estimated pattern for ``non-backdoor'' class pairs may also have a small norm). It maybe difficult to collect such many images for each class, and the associated computational complexity may be quite high.

\vspace{-0.1in}
\section{Method}
\label{sec:method}
\vspace{-0.1in}

Here, we present our {\it efficient} (both computationally and sample-wise) Lagrangian-based RED (L-RED) against {\it imperceptible} BAs with arbitrary number of source classes. Like existing REDs, L-RED contains a pattern estimation step followed by an anomaly detection step.

\vspace{-0.1in}
\subsection{Pattern Estimation}
\vspace{-0.05in}

For {\it efficiency}, we perform pattern estimation for each putative target class $t\in{\mathcal C}$ using all images from $\cup_{s\neq t}{\mathcal D}_s$. Supposing $N$ clean images per class for detection, we perform $K$ pattern estimations, each using $(K-1)N$ clean images. While \cite{Post-ICASSP} performs $K(K-1)$ pattern estimations, each using $N$ clean images for $N$ sufficiently large to ensure detection accuracy, we only need $(K-1)N$ to be sufficiently large to have a significant outlier for ``backdoor'' target class. Thus, {\it the minimum $N$ (and computation) required for our method to accurately detect BAs in practice will be much smaller than for \cite{Post-ICASSP} (especially when $K$ is large).} However, the defender is not aware of the presence of any BAs, the number of source classes, or which they are a priori. Instead of unreasonably assuming ${\mathcal S}^{\ast}\cup t^{\ast}={\mathcal C}$ for BAs, we consider that if there is a BA, it may involve arbitrary number of source classes and any non-target class may possibly be one of the source classes. Correspondingly, we propose the following weighted pattern estimation problem for each putative $t\in{\mathcal C}$:
\vspace{-0.05in}
\begin{equation}\label{eq:opt_raw_imperceptible_efficient}
\vspace{-0.05in}
\begin{aligned}
& \underset{{\bf v}, {\bf w}}{\text{minimize}}
& & ||{\bf v}||_2\\
\vspace{-0.05in}
& \text{subject to}
& & Q_{t}({\bf v}, {\bf w})\triangleq \sum_{s\neq t} w_s q_{st}({\bf v}) ~\geq~ \pi,\\
\vspace{-0.05in}
& & & \sum_{s\neq t} w_s = 1, \quad {\bf w}\geq {\mathbf 0},
\end{aligned}
\vspace{-0.05in}
\end{equation}
where $q_{st}({\bf v}) \triangleq \frac{1}{|\mathcal{D}_s|}\sum_{{\bf x}\in\mathcal{D}_s}{\mathds 1} (f(m({\bf x};{\bf v});\theta)=t)$ is the misclassification fraction from $s$ to $t$ given pattern ${\bf v}$. By solving (\ref{eq:opt_raw_imperceptible_efficient}) for both the weights ${\bf w}$ and ${\bf v}$, for the true target class when there is a BA, we expect the true source classes to be assigned much higher weights than other classes.

Following \cite{DeepFool, Post-ICASSP}, we propose an iterative algorithm to solve (\ref{eq:opt_raw_imperceptible_efficient}). We sequentially update ${\bf v}$ and ${\bf w}$ until $Q_{t}({\bf v}, {\bf w})\geq\pi$, with: initial ${\bf w}$ {\em uniform} in all entries except $w_t=0$ and initial ${\bf v}={\mathbf 0}$ (so that initially $Q_{t}({\bf v}, {\bf w}) \approx 0$).

{\bf Update ${\bf v}$:} In iteration $\tau+1$, given ${\bf v}^{(\tau)}$ and ${\bf w}^{(\tau)}$ from iteration $\tau$, we update ${\bf v}$ by ${\bf v}^{(\tau+1)}={\bf v}^{(\tau)}+\Delta{\bf v}^{(\tau+1)}$, where
\begin{equation}\label{eq:opt_update_v}
\begin{aligned}
\Delta{\bf v}^{(\tau+1)} = \underset{||{\bf z}||_2=\delta}{\text{argmax}} ~ Q_{t}({\bf v}^{(\tau)} + {\bf z}, {\bf w}^{(\tau)})
\end{aligned}
\end{equation}
with $\delta$ a small step size. Since $Q_{t}({\bf v}, {\bf w})$ is not differentiable in ${\bf v}$ due to the indicator function, we find an approximate solution to (\ref{eq:opt_update_v}) by just one gradient ascent step on a differentiable surrogate of $Q_t(\cdot, \cdot)$:
\begin{equation}\label{eq:delta_v}
\begin{aligned}
\Delta{\bf v}^{(\tau+1)} \approx \delta \nabla_{\bf v} J_{t}({\bf v}^{(\tau)}, {\bf w}^{(\tau)}) / ||\nabla_{\bf v} J_{t}({\bf v}^{(\tau)}, {\bf w}^{(\tau)})||_2
\end{aligned}
\end{equation}
where $J_{t}({\bf v}, {\bf w}) \triangleq \sum_{s\neq t}w_s \frac{1}{|\mathcal{D}_s|}\sum_{{\bf x}\in\mathcal{D}_s} \log{p(t|m({\bf x};{\bf v});\theta)}$ and $p(c|{\bf x}; \theta)$ is the classifier's posterior for class $c\in{\mathcal C}$ for any image ${\bf x}\in{\mathcal X}$.

{\bf Update ${\bf w}$:} We aim to have the source classes to be assigned higher weights. In iteration $\tau+1$, one may naively hard-assign $w_c=1$ to class $c=\argmax_{s\neq t}q_{st}({\bf v}^{(\tau+1)})$ because with a small-norm ${\bf v}$, misclassification fraction from a source class to the true BA target class tends to be larger than for non-source classes. However, after choosing this class, updating ${\bf v}$ will be performed on clean images associated with only this class in all subsequent iterations. As uniform weights are initially assigned to all classes except $t$, in early iterations, updating ${\bf v}$ may cause any class $c\notin{\mathcal S}^{\ast}$ to temporally have the largest $q_{ct}({\bf v})$. In this case, with hard-assignment, subsequent pattern estimations will be equivalently performed for a ``non-backdoor'' class pair -- a poor local optima with a large $||{\bf v}||_2$ will be obtained. Thus, we propose the following Lagrangian to update ${\bf w}$ in iteration $\tau+1$ subject to a specified level of ``randomness'' (entropy of ${\bf w}$):
\begin{equation}\label{eq:opt_annealing_imperceptible}
\begin{aligned}
& \underset{{\bf w}}{\text{maximize}}
& & Q_{t}({\bf v}^{(\tau+1)}, {\bf w}) - T^{(\tau+1)} \sum_{s\neq t} w_s \log{w_s}\\
\vspace{-0.05in}
& \text{subject to}
& & \sum_{s\neq t} w_s = 1,
\end{aligned}
\end{equation}
where the Lagrange multiplier $T^{(\tau+1)}\geq0$ can be viewed as the current iteration's ``temperature''. Note that if $T^{(\tau+1)}=0$, solving (\ref{eq:opt_annealing_imperceptible}) is equivalent to hard weight assignment; if $T^{(\tau+1)}$ is large, solving (\ref{eq:opt_annealing_imperceptible}) is equivalent to max-entropy given $w_t=0$. To avoid poor local optima caused by hard weight assignment, we need $T$ to be large when $Q_{t}({\bf v}, {\bf w})$ is small in early iterations. As $Q_{t}({\bf v}, {\bf w})$ grows large, $T$ should {\it gradually} decrease, such that higher weights are {\it gradually} assigned to classes with large $q_{\cdot t}({\bf v})$ -- these classes are likely the source classes if $t$ is the true BA target class. We choose to schedule $T$ automatically as:
\begin{equation}\label{eq:update_T}
\begin{aligned}
T^{(\tau+1)} = -\log{Q_{t}({\bf v}^{(\tau+1)}, {\bf w}^{(\tau)})} \in (0, +\infty)
\end{aligned}
\end{equation}
for $Q_{t}({\bf v}^{(\tau+1)}, {\bf w}^{(\tau)})>0$ (otherwise, ${\bf w}$ is set to uniform with $w_t=0$). Note that $T\neq0$ since the algorithm is terminated when $Q_{t}({\bf v}, {\bf w})\geq\pi$ for some $\pi<1$. Given this scheduled $T^{(\tau+1)}$, the closed-form solution to (\ref{eq:opt_annealing_imperceptible}) is:
\begin{equation}\label{eq:update_w}
\begin{aligned}
w_s^{(\tau+1)} = \frac{\exp{[q_{st}({\bf v}^{(\tau+1)})/T^{(\tau+1)}]}}{\sum_{c\neq t}\exp{[q_{ct}({\bf v}^{(\tau+1)})/T^{(\tau+1)}]}}, ~\forall s\neq t
\end{aligned}
\vspace{-0.05in}
\end{equation}
with the derivation in Appx. \ref{apdx:updating_w}. 
Pseudocode for our pattern estimation algorithm is in Appx. \ref{apdx:algotithm}.

\vspace{-0.1in}
\subsection{Anomaly Detection (AD)}
\vspace{-0.05in}

AD of L-RED uses the similar hypothesis test idea as the approach in \cite{Post-ICASSP}, but is much simpler in form. We first obtain a detection statistic for each putative target class $t$ as $r_t=(||{\bf v}_t||_2)^{-1}$, where ${\bf v}_t$ is the {\it estimated pattern} for class $t$. To test the null hypothesis that there is no attack, we fit a null distribution $G(\cdot)$ using all statistics except $r_{\rm max}=\max_t r_t$ (associated with the class with the smallest estimated pattern norm). We choose $G(\cdot)$ in form of Gamma distribution, while other single tailed distribution form should also work \cite{Post-ICASSP}. Then we calculate the order statistic p-value for $r_{\rm max}$ under $G(\cdot)$ as ${\rm pv}=1-G(r_{\rm max})^K$. A detection threshold $\phi$ is chosen, such that the null hypothesis is rejected (i.e. an attack is detected) with confidence $(1-\phi)$ if ${\rm pv} < \phi$. When an attack is detected, $\hat{t}=\argmax_t r_t$ is inferred as the target class.

\vspace{-0.1in}
\section{Experiments}
\label{sec:experiments}
\vspace{-0.1in}

We compare the {\it effectiveness} and {\it efficiency} of L-RED with existing REDs against BAs on CIFAR-10 (with 50k training images and 10k test images, both evenly distributed in 10 classes). Key results for other datasets are also shown.

\vspace{-0.1in}
\subsection{Attacker: Creating Backdoor Training Images}
\label{subsec:exp_attacker}
\vspace{-0.05in}

We consider a ``global'' pattern with maximum perturbation size $2/255$ (see Fig. 5 of \cite{Haoti}) and a ``local'' square pattern which perturbs only four pixels of an image by $50/255$. For {\it each} pattern, using the standard backdoor training image crafting approach in Sec. \ref{subsec:backdoor}, we create three groups of attacks (10 attacks per group) with {\it one} source class, {\it three} source classes, and {\it nine} source classes, respectively. The backdoor pattern generation, our (arbitrary) choices of source classes and target class, and the number of backdoor training images are detailed in Appendix \ref{apdx:details_attack}.

\vspace{-0.1in}
\subsection{Trainer: Training on Poisoned Training Set}
\label{subsec:exp_trainer}
\vspace{-0.05in}

Here, we use the ResNet-18 \cite{ResNet} DNN architecture. For each attack, one classifier is trained on the poisoned training set. Training is performed for 100 epochs with batch size 32 and learning rate $10^{-3}$. Data augmentation including random cropping and random horizontal flipping are used during training. For convenience, we name the six groups of classifiers being attacked as ``G-1'', ``G-3'', ``G-9'', ``L-1'', ``L-3'', and ``L-9'', respectively, where ``G'' represents ``global'', ``L'' represents ``local'' (backdoor pattern), and the numbers represent the number of source classes. We also train 10 classifiers on the clean CIFAR-10 training set as the control group (named ``C-0'') for false detection rate evaluation.

\begin{table}[t]
	\begin{center}
		\caption{Attack success rate (ASR) and clean test accuracy (CTA) for the six group of classifiers being attacked.}
		\resizebox{0.48\textwidth}{!}{
			\begin{tabular}{ |c|c|c|c|c|c|c| }
				\hline
				& G-1 & G-3 & G-9 & L-1 & L-3 & L-9\\
				\hline
				ASR & $92.1\pm1.9$ & $94.1\pm1.1$ & $96.6\pm0.3$ & $93.1\pm1.6$ & $91.1\pm0.8$ & $91.9\pm0.7$\\
				\hline 
				CTA & $93.6\pm0.2$ & $93.5\pm0.2$ & $93.5\pm0.2$ & $93.7\pm0.1$ & $93.6\pm0.2$ & $93.6\pm0.2$\\
				\hline
			\end{tabular}\label{tab:ASR_CTA}}
	\end{center}
\vspace{-0.2in}
\end{table}

{\it From the experimenter's perspective}, we show the effectiveness of the attacks through attack success rate (ASR) and clean test accuracy (CTA), using the test set of CIFAR-10. ASR of an attack is defined as the fraction of test images from the source class(es) being classified to the target class when the backdoor pattern is embedded. In Tab. \ref{tab:ASR_CTA}, for all attacks, the ASRs are uniformly high, with almost no degradation in CTA compared with group C-0 (with CTA $93.8\pm0.1$) -- the attacks are all successful.

\begin{table}[t]
	\begin{center}
		\caption{Detection accuracy (fraction of successful detection) of the defenses on the seven groups of classifiers.}
		\resizebox{0.40\textwidth}{!}{
			\begin{tabular}{ |c|c|c|c|c|c|c|c| }
				\hline
				& G-1 & G-3 & G-9 & L-1 & L-3 & L-9 & C-0\\
				\hline
				P-RED & 5/10 & 9/10 & 7/10 & 2/10 & 3/10 & 3/10 & 7/10\\
				\hline 
				U-RED & 2/10 & 8/10 & {\bf 10/10} & 2/10 & 5/10 & {\bf 10/10} & {\bf 10/10}\\
				\hline
				L-RED & {\bf 10/10} & {\bf 10/10} & {\bf 10/10} & {\bf 10/10} & {\bf 10/10} & {\bf 10/10} & {\bf 10/10}\\
				\hline
				L-RED' & 9/10 & {\bf 10/10} & {\bf 10/10} & 9/10 & 8/10 & {\bf 10/10} & 9/10\\
				\hline
			\end{tabular}\label{tab:accuracy}}
	\end{center}
	\vspace{-0.2in}
\end{table}

\vspace{-0.125in}
\subsection{Defender: Detecting if Classifier is Attacked}
\label{subsec:exp_defender}
\vspace{-0.05in}

We compare {\bf L-RED} with two other REDs using the six groups of classifiers being attacked and the clean classifiers in C-0:
(1) {\bf P-RED}: the state-of-the-art RED against imperceptible BAs in \cite{Post-ICASSP} that performs pattern estimation for each class pair;
(2) {\bf U-RED}: RED with the ${\mathcal S}^{\ast}\cup t^{\ast}={\mathcal C}$ assumption. Note that U-RED is a special case of L-RED with ${\bf w}$ fixed to uniform.
For all these defenses, we choose $\pi=0.9$, pattern estimation step size $\delta=10^{-4}$, and detection (confidence) threshold $\phi=0.05$, while in general, these choices are not critical to the performance of RED against imperceptible BAs \cite{Post-ICASSP}. For L-RED, P-RED, and U-RED, we use 8 clean images per class for detection. We also consider a ``data-limited'' case, denoted by {\bf L-RED'}, where only 2 clean images per class are available. We reasonably assume that all clean images for detection are correctly classified. For each classifier to be inspected, the clean images for detection are randomly sampled from the test set of CIFAR-10, which are independent from the training (and attack crafting) process.\\
{\bf Accuracy Evaluation:} For all the defenses to be evaluated, a successful detection of an attack requires also a correct inference of the target class. For the clean classifiers in C-0, ``no attack detected'' is deemed a successful detection. In Tab. \ref{tab:accuracy}, we show the fraction of successful detection for all the defenses for the seven groups of classifiers. For U-RED, although perfect detection is achieved for G-9 and L-9 where the ${\mathcal S}^{\ast}\cup t^{\ast}={\mathcal C}$ assumption is satisfied, with no false detection on C-0, the detection accuracy on G-1, G-3, L-1, and L-3 (where $|{\mathcal S}^{\ast}| < |{\mathcal C}|$) is poor. P-RED shows limited detection capability with 8 clean images per class for attacks with the ``global'' pattern, which is consistent with Fig. 13 in \cite{Post-ICASSP}, but it fails for most of the attacks with the ``local'' pattern. In comparison, L-RED achieves perfect detection of all the attacks, {\it regardless of the number of source classes}, and with no false detection. Notably, with only 2 clean images per class, L-RED' detects 56 out of 60 attacks with only 1 out of 10 false detection, which is generally better result than both P-RED and U-RED. Additional results (for e.g. weight assignment) and estimated backdoor patterns of L-RED are in Appx. \ref{apdx:insights} and Appx. \ref{apdx:estimated_pattern}, respectively.\\
{\bf Efficiency Evaluation:} As discussed below (\ref{eq:opt_raw_imperceptible_efficient}) and demonstrated experimentally, to achieve high detection accuracy, defenses like P-RED need a ``sufficiently large'' number of clean images per class (i.e. a large $N$), while L-RED does not. As shown in Tab. \ref{tab:time}, L-RED and P-RED need similar average execution time for each of the seven groups of classifiers. However, L-RED' with $N=2$ exhibits a much lower time consumption than both L-RED and P-RED with $N=8$. The execution time for U-RED is similar as for L-RED, hence is not shown here due to page limitations. All experiments above are performed on a RTX2080-Ti (11GB) GPU.

\begin{table}[t]
	\begin{center}
		\caption{Average execution time (in seconds) of the defenses on the seven groups of classifiers.}
		\resizebox{0.32\textwidth}{!}{
			\begin{tabular}{ |c|c|c|c|c|c|c|c| }
				\hline
				& G-1 & G-3 & G-9 & L-1 & L-3 & L-9 & C-0\\
				\hline
				P-RED & 480 & 473 & 531 & 478 & 479 & 394 & 541\\
				\hline
				L-RED & 562 & 615 & 636 & 529 & 482 & 536 & 535\\
				\hline
				L-RED' & 136 & 161 & 167 & 129 & 169 & 163 & 157\\
				\hline
			\end{tabular}\label{tab:time}}
	\end{center}
\vspace{-0.2in}
\end{table}

\begin{table}[t]
	\begin{center}
		\caption{Order statistic p-values for both clean and attacked classifiers on MNIST, FMNIST, GTSRB, and CIFAR-100, when applying our defense (``u.f.'' for ``underflow'').}
		\resizebox{0.40\textwidth}{!}{
			\begin{tabular}{ |c|c|c|c|c| }
				\hline
				& MNIST & FMNIST & GTSRB & CIFAR-100\\
				\hline
				Attacked & u.f. & $8.44\times10^{-7}$ & $3.41\times10^{-9}$ & u.f.\\
				\hline
				Clean & 0.300 & 0.331 & 0.152 & 0.551\\
				\hline
			\end{tabular}\label{tab:pvalues}}
	\end{center}
	\vspace{-0.2in}
\end{table}

\vspace{-0.1in}
\subsection{Experiments on Other Datasets}
\vspace{-0.05in}

We create one attack on each of the MNIST, FMNIST, GTSRB, and CIFAR-100 datasets. For each dataset, one classifier is trained on the backdoor poisoned dataset and one clean classifier is trained without attack. Details of the attacks and training are shown in Appx. \ref{apdx:other_datasets}. We apply L-RED with the same settings as above to inspect these classifiers, except that for classifiers trained on GTSRB and CIFAR-100, we use {\it only one} clean image per class for detection. In Tab. \ref{tab:pvalues}, we show the order statistic p-values obtained for each classifier by our defense. All attacks are detected with no false detection of any clean classifiers using threshold $\phi=0.05$. For BA on CIFAR-100, our defense requires less than 3 hours for detection while the method in \cite{Post-ICASSP} requires several days.

\vspace{-0.1in}
\section{Conclusion}
\label{sec:conclusion}
\vspace{-0.1in}

We proposed an Lagrangian-based RED against imperceptible BAs with arbitrary number of source classes. Our defense requires very few clean images per class for detection and is computationally efficient.

\vfill\pagebreak

\bibliographystyle{IEEEbib}
\bibliography{refs}

\vfill\pagebreak

\appendix
\section{Appendix}

\subsection{Derivation of the Updating Rule for the Weight Vector}
\label{apdx:updating_w}

Here we neglect the iteration indices for simplicity. With ${\bf v}$ fixed and $T$ scheduled (using Eq. (\ref{eq:update_T})) based on the current iterations' weighted misclassification fraction $Q_{t}({\bf v}, {\bf w})$, solving (\ref{eq:opt_annealing_imperceptible}) over ${\bf w}$ yields a closed-form solution, which is given as (\ref{eq:update_w}). Here, we show the details.

We first cast (\ref{eq:opt_annealing_imperceptible}) into the following Lagrangian:
\begin{equation}
\begin{aligned}
L({\bf w}, \nu) = \sum_{s\neq t} w_s q_{st}({\bf v}) - T \sum_{s\neq t} w_s \log{w_s} + \nu (\sum_{s\neq t} w_s - 1).
\end{aligned}
\end{equation}
For all $s\neq t$, we obtain the following partial derivative:
\begin{equation}
\begin{aligned}
\frac{\partial L}{\partial w_s} = q_{st}({\bf v}) - T(\log{w_s} + 1) + \nu.
\end{aligned}
\end{equation}
By setting the above partial derivative to 0, we obtain, for all $s\neq t$:
\begin{equation}
\begin{aligned}
w_s = \exp(\frac{q_{st}({\bf v}) + \nu - T}{T})
\end{aligned}
\end{equation}
By plugging the above into the constraint $\sum_{s\neq t}w_s=1$, we get the closed-form solution (\ref{eq:update_w}).

\subsection{L-RED Pattern Estimation Algorithm}
\label{apdx:algotithm}

\begin{algorithm}
	\caption{L-RED pattern estimation.}\label{alg:RE_imperceptible}
	\begin{algorithmic}[1]
		\State {\bf Inputs}: putative target class $t$, clean images $\cup_{s\neq t}{\mathcal D}_s$, classifier $f(\cdot;\theta)$, step size $\delta$, target misclassification fraction $\pi$, maximum number of iterations $\tau_{\rm max}$
		\State {\bf Initialization}: ${\bf v}^{(0)}={\mathbf 0}$, $w_s^{(0)}=1/(K-1)$ for $\forall s\neq t$, $w_t^{(0)}=0$, $\tau=0$, estimated pattern ${\bf v}_t={\mathbf 0}$
		\While{$\tau < \tau_{\rm max}$}
		\State ${\bf v}_t = {\bf v}^{(\tau+1)} = {\bf v}^{(\tau)} + \delta \frac{\nabla_{\bf v} J_{t}({\bf v}^{(\tau)}, {\bf w}^{(\tau)})}{||\nabla_{\bf v} J_{t}({\bf v}^{(\tau)}, {\bf w}^{(\tau)})||_2}$
		\If{$Q_{t}({\bf v}^{(\tau+1)}, {\bf w}^{(\tau)}) \geq \pi$}
		\State {\bf break}
		\ElsIf{$Q_{t}({\bf v}^{(\tau+1)}, {\bf w}^{(\tau)}) > 0$}
		\State $T^{(\tau+1)} = -\log{Q_{t}({\bf v}^{(\tau+1)}, {\bf w}^{(\tau)})}$
		\ForAll{$s\neq t$}
		\State $w_s^{(\tau+1)} = \frac{\exp{[q_{st}({\bf v}^{(\tau+1)})/T^{(\tau+1)}]}}{\sum_{c\neq t}\exp{[q_{ct}({\bf v}^{(\tau+1)})/T^{(\tau+1)}]}}$
		\EndFor
		\Else
		\State ${\bf w}^{(\tau+1)} = {\bf w}^{(0)}$
		\EndIf
		\State $\tau\texttt{++}$
		\EndWhile
		\State {\bf Outputs}: ${\bf v}_t$
	\end{algorithmic}
\end{algorithm}

In early stage of this work, we also attempted to solve (\ref{eq:opt_raw_imperceptible_efficient}) by solving the following surrogate problem:
\begin{equation}\label{eq:opt_gradeitn}
\begin{aligned}
& \underset{{\bf v}, {\bf w}}{\text{maximize}}
& & J_{t}({\bf v}, {\bf w})\\
\vspace{-0.05in}
& \text{subject to}
& & \sum_{s\neq t} w_s = 1, \quad {\bf w}\geq {\mathbf 0},
\end{aligned}
\end{equation}
using gradient ascent with the same initialization as for Algorithm \ref{alg:RE_imperceptible} and proper projection of ${\bf w}$, until $Q_{t}({\bf v}, {\bf w}) \geq \pi$ is satisfied. However, the performance of this approach is very sensitive to the choice of the step size for updating ${\bf v}$ and ${\bf w}$, and automatic scheduling of the step size is almost infeasible in practice (when the defender does not know if an attack exists). If the step size is small, ${\bf w}$ will be close to uniform when $Q_{t}({\bf v}, {\bf w}) \geq \pi$ is satisfied -- the norm of the estimated pattern for the BA target class will not be abnormally small (for the case where $|{\mathcal S}^{\ast}|\ll K$). If the step size is large, non-source classes may be assigned overly high weights, which hurts detection accuracy. Due to these issues, L-RED is a more reliable choice for practical defenders.

\subsection{Potential Extension to the Perceptible Case}
\label{apdx:perceptible}

We focused on detecting imperceptible BAs with embedding mechanism Eq. (\ref{eq:bd_imperceptible}) (which is the only effective existing way of imperceptible backdoor embedding to our knowledge). However, backdoor patterns can also be perceptible but seemingly innocuous to the scene, e.g. a pair of glasses on a face \cite{Targeted}. Perceptible backdoor patterns are embedded in a clean image ${\bf x}\in{\mathcal X}$ by:
\begin{equation}\label{eq:bd_perceptible}
m_{\rm p}({\bf x}; \{{\bf u}^{\ast}, {\bf m}^{\ast}\}) = {\bf x}\odot({\mathbf 1} - {\bf m}^{\ast})+{\bf u}^{\ast}\odot{\bf m}^{\ast}
\end{equation}
where ${\bf u}^{\ast}$ is an image-wide pattern (usually with small $||{\bf u}^{\ast}||_0$ to represent the backdoor ``object'', e.g. the pair of glasses); ${\bf m}^{\ast}$ is an image-wide mask with $m_{i,j}\in\{0, 1\}$\footnote{A blended embedding mechanism is mentioned by \cite{Targeted} with $m_{i,j}\in[0, 1]$.}; $\odot$ represents element-wise multiplication, which applies ${\bf m}^{\ast}$ to all colors/channels of ${\bf x}$ (\cite{BadNet, NC}). Note that ${\bf m}^{\ast}$ is usually associated with the pattern ${\bf u}^{\ast}$, such that $m_{i, j}=1$ only if $u_{i, j, k}>0$ for any channel $k$, hence both $||{\bf m}^{\ast}||_0$ and $||{\bf m}^{\ast}||_1$ are typically small \cite{NC}.

REDs detecting perceptible BAs rely on a similar idea with REDs for the imperceptible case: there is a small $l_1$ norm mask (and a pattern restricted by this mask) that induces a high misclassification fraction from the BA's source classes to the BA's target class. RED proposed by \cite{NC} which relies on the assumption ${\mathcal S}^{\ast}\cup t^{\ast}={\mathcal C}$ solves the following problem for each putative target class $t$:
\begin{equation}\label{eq:opt_raw_perceptible}
\begin{aligned}
& \underset{{\bf u}, {\bf m}}{\text{minimize}}
& & ||{\bf m}||_1\\
& \text{subject to}
& & \frac{1}{K-1}\sum_{s\neq t}\frac{1}{|{\mathcal D}_s|}\sum_{{\bf x}\in{\mathcal D}_s}{\mathds 1} (f(m_{\rm p}({\bf x};\{{\bf u}, {\bf m}\});\theta)=t)\\
& & & ~\geq~ \pi,
\end{aligned}
\end{equation}
where $K = |\mathcal{C}|$, and hopes to find a mask with an abnormally small $l_1$ norm for the backdoor target class when there is an attack. To detect perceptible BAs with arbitrary number of source classes {\it without considering the efficiency}, we can easily formulate a (pattern, mask) estimation problem for each class pair $(s, t)\in\mathcal{C}\times\mathcal{C}$ ($s\neq t$) based on (\ref{eq:opt_raw_perceptible}):
\begin{equation}\label{eq:opt_raw_perceptible_cp}
\begin{aligned}
& \underset{{\bf u}, {\bf m}}{\text{minimize}}
& & ||{\bf m}||_1\\
& \text{subject to}
& & \frac{1}{|\mathcal{D}_s|}\sum_{{\bf x}\in\mathcal{D}_s}{\mathds 1} (f(m_P({\bf x};\{{\bf u}, {\bf m}\});\theta)=t) ~\geq~ \pi.
\end{aligned}
\end{equation}
Note that (\ref{eq:opt_raw_perceptible_cp}) has a very similar form with (\ref{eq:opt_raw_imperceptible}). We believe that our efficient L-RED can be extended for the perceptible case by introducing a weight vector to (\ref{eq:opt_raw_perceptible_cp}) to formulate a problem similar to (\ref{eq:opt_raw_imperceptible_efficient}). This is a promising future research direction based on the current work.

\subsection{Attack Crafting Details}
\label{apdx:details_attack}

In our experiments, pixel values are always normalized to $[0, 1]$. We also consider practical case by using 8-bit finite precision for pixel values, such that a valid pixel value will be in the set $\{0, 1/255, \cdots, 1\}$. 

\subsubsection{Backdoor Patterns}

\begin{figure}[t]
	\centering
	\begin{subfigure}[b]{.45\linewidth}
		\centering
		\centerline{\includegraphics[width=\linewidth]{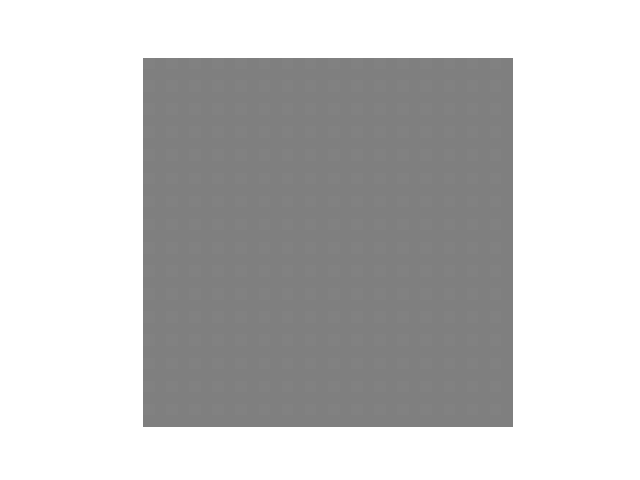}}
		\subcaption{``global'' pattern}
	\end{subfigure}
	\begin{subfigure}[b]{.45\linewidth}
		\centering
		\centerline{\includegraphics[width=\linewidth]{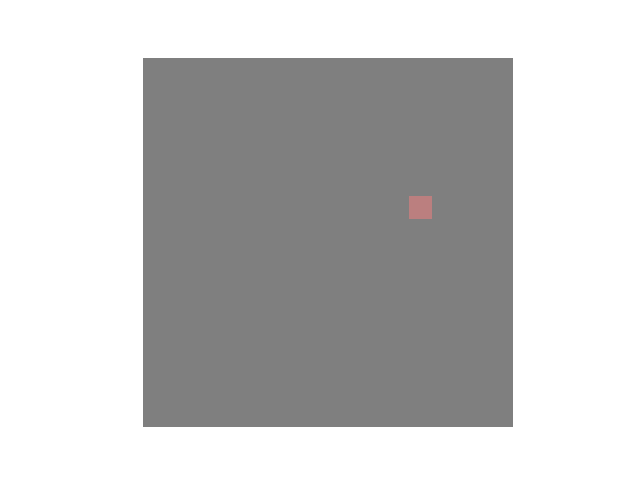}}
		\subcaption{``local'' pattern}
	\end{subfigure}
	\begin{subfigure}[b]{.45\linewidth}
		\centering
		\centerline{\includegraphics[width=\linewidth]{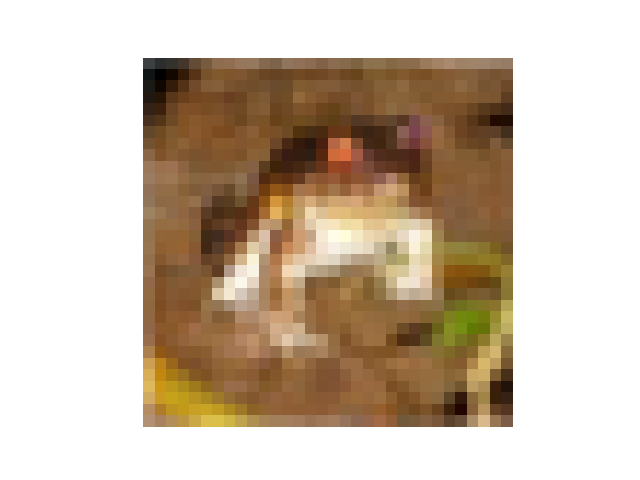}}
		\subcaption{with ``global'' pattern}
	\end{subfigure}
	\begin{subfigure}[b]{.45\linewidth}
		\centering
		\centerline{\includegraphics[width=\linewidth]{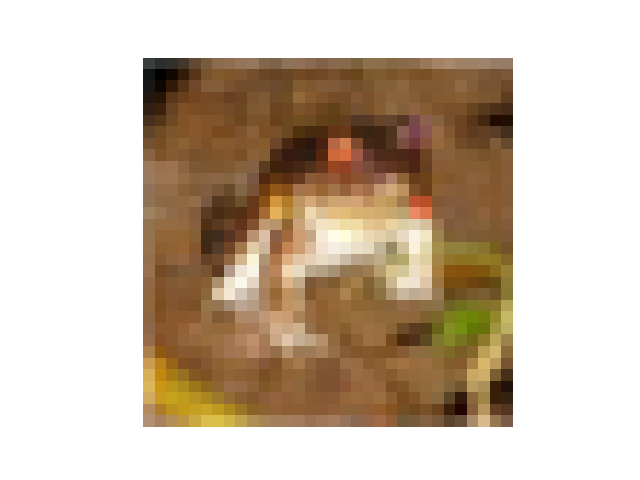}}
		\subcaption{with ``local'' pattern}
	\end{subfigure}
	\begin{subfigure}[b]{.45\linewidth}
		\centering
		\centerline{\includegraphics[width=\linewidth]{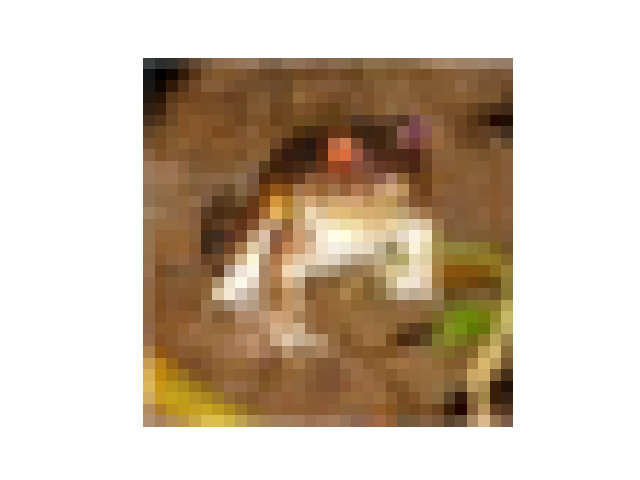}}
		\subcaption{clean}
	\end{subfigure}
	\caption{Example backdoor patterns (with 127/255 offset for better visualization), example backdoor training images embedded with each backdoor pattern, and originally clean image.}
	\label{fig:bd_example}
\end{figure}

In our experiments on CIFAR-10, we considered a ``global'' pattern and a ``local'' pattern. The global pattern was first considered in \cite{Haoti} (see their Fig. 5). For a $2\times2$ window applied to any spatial location in an image, there is one and only one pixel being perturbed (in all color channels). Here, we set the perturbation size to be $2/255$ for pixels and channels being perturbed. The local square pattern is similar to the local patterns used in \cite{SS} and \cite{DataLimited}. For each attack, we randomly select a $2\times2$ square and a color channel, and perturb the four pixels in the selected channel by $50/255$. Example backdoor patterns and backdoor training images are shown in Fig. \ref{fig:bd_example}. Note that the embedded backdoor pattern are barely noticeable to humans.

\subsubsection{Source Classes and Target Class}

For simplicity, we enumerate the 10 classes of CIFAR-10, i.e., `plane', `car', `bird', `cat', `deer', `dog', `frog', `horse', `ship', and `truck', as class `1-10'. The target classes for attacks using the ``global'' pattern and the ``local'' pattern are class 10 and class 2, respectively. For the ``global'' pattern, the source classes for attacks with one source class and attacks with three source classes are class 1 and classes $\{1, 7, 9\}$, respectively. For the ``local'' pattern, the source classes for attacks with one source class and attacks with three source classes are class 1 and classes $\{1, 5, 6\}$, respectively. For both backdoor patterns, attacks with nine source classes use all the classes except for the target class as source classes, i.e., following the ${\mathcal S}^{\ast}\cup t^{\ast}={\mathcal C}$ assumption. All the above choices of source classes and target class are arbitrary.

\subsubsection{Number of Backdoor Training Images}

In general, the ``global'' pattern is ``easier'' to be learned than the ``local'' pattern. Hence, fewer backdoor training images are required if the ``global'' pattern is used to launch a BA. For the ``global'' pattern, for attacks with one source class, three source classes, and nine source classes, 150, 50, and 15 back door training images per source class are created, respectively. For the ``local'' pattern, for attacks with one source class, three source classes, and nine source classes, 900, 300, and 100 back door training images per source class are created, respectively. Note that in practice, an attacker can either poison the training set with more backdoor training images, or increase the perturbation size (or the number of pixels to be perturbed for any ``local'' patterns) to achieve a better attack effectiveness.

\subsubsection{Emulation of Practical Backdoor Poisoning}

Recall from Sec. \ref{subsec:backdoor} that backdoor training images are created using clean images from the source classes, embedded with the backdoor pattern, and labeled to the target class. In our experiments on CIFAR-10, the originally clean images used to generate the backdoor training images are from the training set of CIFAR-10. To emulate the practical case where those clean images are collected by the attacker but not possessed by the trainer, we remove those clean images used for generating backdoor training images from the CIFAR-10 training set. Then the {\it poisoned} CIFAR-10 training set still contains 50k training images.

\subsection{Insights behind the Accuracy Evaluation Results}
\label{apdx:insights}

We showed that L-RED and L-RED' outperform existing REDs on detecting imperceptible BAs. In particular, L-RED achieves perfect detection results with no false detection. The main reasons behind the success of our defense are explained in Sec. \ref{sec:method} as the foundation of our pattern estimation algorithm (Algorithm \ref{alg:RE_imperceptible}). We jointly estimate a pattern and a weight vector. While updating the weights, we introduce an entropy term with a gradually decreasing Lagrange multiplier (temperature parameter), such that more weights are gradually assigned to classes that are more likely the source classes. Here, we visualize such weight assignment by first introducing the concept of ``collateral damage'' which is first observed by \cite{Post-ICASSP}.

\begin{definition}
A classifier undergoing a BA with source classes ${\mathcal S}^{\ast}$ and target class $t^{\ast}$ (${\mathcal S}^{\ast}\cup t^{\ast}\neq {\mathcal C}$) is said to suffer from {\bf collateral damage} if for any class $\tilde{s}\notin{\mathcal S}^{\ast}\cup t^{\ast}$, the test-time group misclassification from $\tilde{s}$ to $t^{\ast}$ induced by the same backdoor pattern is significantly higher than the class confusion from $\tilde{s}$ to $t^{\ast}$ for clean test images. Moreover, the higher the group misclassification from $\tilde{s}$ to $t^{\ast}$ is, the more severe the collateral damage is for class $\tilde{s}$.
\end{definition}

The possible reason for the collateral damage is that the backdoor pattern may be learned independently from the source classes, such that a test image not from any source classes may be classified to the target class when the backdoor pattern exists.

\begin{figure}[t]
	\centering
	\begin{subfigure}[b]{.95\linewidth}
		\centering
		\centerline{\includegraphics[width=\linewidth]{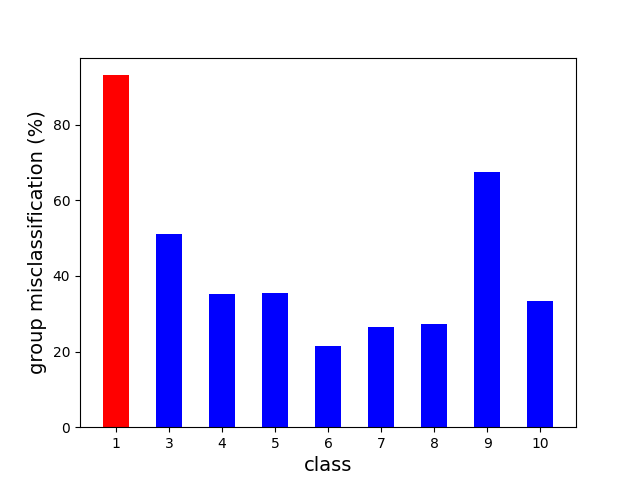}}
		\subcaption{collateral damage}\label{fig:weight_assignment_collateral_damage}
	\end{subfigure}
	\begin{subfigure}[b]{.95\linewidth}
		\centering
		\centerline{\includegraphics[width=\linewidth]{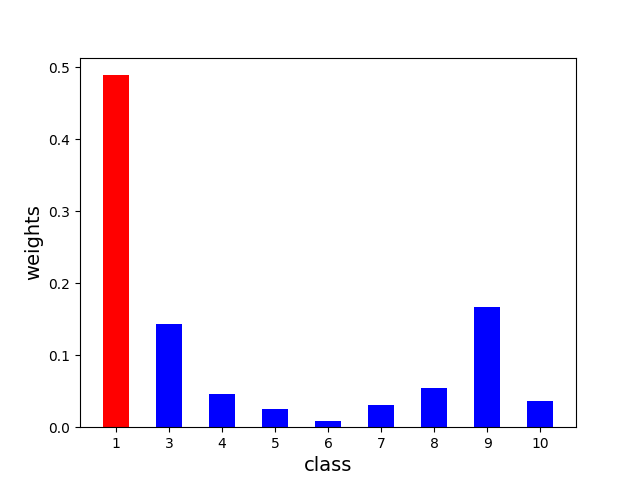}}
		\subcaption{weights}\label{fig:weight_assignment_weights}
	\end{subfigure}
	\caption{Group misclassification fraction and weight assignment (when using L-RED) for all non-target classes, averaged over the ten classifiers from the L-1 group.}
	\label{fig:weight_assignment}
\end{figure}

For L-RED, if there is an attack, for the true backdoor target class, {\it we expect that when the stopping condition of Algorithm \ref{alg:RE_imperceptible} is met, weights will be assigned mainly to the backdoor source classes and classes suffering severe collateral damage (if there are any)}. Here, we consider the ten classifiers from the L-1 group, where the source class and the target class are class 1 and class 2, respectively. For each of the ten classifiers, for each class except the target class, we embed the same backdoor pattern used by the attacker into the clean test images (that are not involved in classifier's training) labeled to this class to evaluate the group misclassification fraction to the target class. The average group misclassification fraction for each non-target class over the ten classifiers is shown in Fig. \ref{fig:weight_assignment_collateral_damage}. For each of the ten classifiers, we also consider the weight assignment at the end of pattern estimation for the true backdoor target class when L-RED is applied. The weight for each class is also averaged over the ten classifiers and is shown in Fig. \ref{fig:weight_assignment_weights}. As we have expected, on average, the true source class, class 1, is assigned a much larger weight than all other classes. Class 3 and class 9 are also assigned some weights since they suffer more sever collateral damage than other non-source classes.

\subsection{Estimated Backdoor Patterns}
\label{apdx:estimated_pattern}

\begin{figure}[t]
	\centering
	\begin{subfigure}[b]{.45\linewidth}
		\centering
		\centerline{\includegraphics[width=\linewidth]{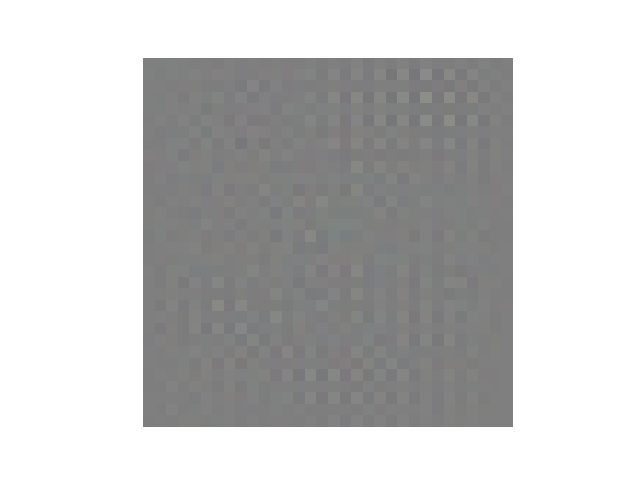}}
		\subcaption{estimated ``global'' pattern}\label{fig:pattern_estimation_additional_static}
	\end{subfigure}
	\begin{subfigure}[b]{.45\linewidth}
		\centering
		\centerline{\includegraphics[width=\linewidth]{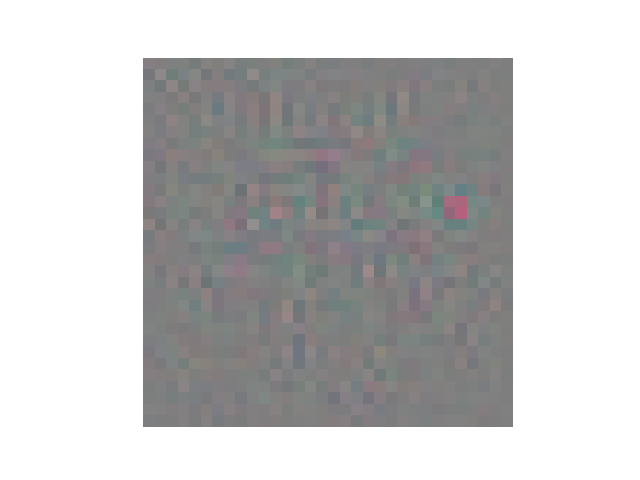}}
		\subcaption{estimated ``local'' pattern}\label{fig:pattern_estimation_main_square}
	\end{subfigure}
	\caption{Example backdoor patterns (with 127/255 offset to elucidate negative perturbations) estimated by L-RED for the two attacks whose true backdoor patterns being used are shown in Fig. \ref{fig:bd_example}.}
	\label{fig:pattern_estimation_main}
\end{figure}

In the main paper, we show that L-RED successfully detects all the attacks with no false detections. In Fig. \ref{fig:pattern_estimation_main}, we show example backdoor patterns estimated by L-RED. In particular, we focus on the two attacks whose backdoor patterns are shown in Fig. \ref{fig:bd_example}. The estimated patterns clearly contain the main features of the true backdoor patterns being used (see the recurrent pattern patches in the left figure and the red square in the right figure).

\subsection{Details of Experiments on Other Datasets}
\label{apdx:other_datasets}

Here, we provide details for the experiments on MNIST \cite{MNIST}, FMNIST \cite{FashionMNIST}, GTSRB \cite{GTSRB}, and CIFAR-100 \cite{CIFAR10}. MNIST is a handwritten digit image dataset containing 60k training images and 10k test images, both evenly distributed in 10 classes. FMNIST (Fashion-MNIST) also contains 60k training images and 10k test images, both evenly distributed in 10 classes. Images in both MNIST and FMNIST are $28\times28$ gray scale images. GTSRB is a traffic sign dataset with 43 classes, 26640 training images, and 12630 test images. We resize each image in GTSRB to $32\times32$. CIFAR-100 contains 50k training images and 10k test images, both evenly distributed in 10 classes. Similar to the images in CIFAR-10, the images in CIFAR-100 are $32\times32$ colored images. For convenience, the class indexing for these datasets used in this section follows the official description of the datasets.

\subsubsection{Attack Crafting}

For the attack on MNIST, we use a ``local chessboard'' pattern, i.e. a patch of the ``chessboard'' pattern located at the bottom right corner of each image to be embedded in. We choose the patch size to be $6\times6$ and perturbation size to be 25/255. The source classes and the target class are class $\{5, 7, 10\}$ and class 9, respectively. 500 backdoor training images per source class are created to poison the training set.

For the attack on FMNIST, we use a ``single pixel'' perturbation backdoor pattern \cite{SS, CI}, where the pixel being perturbed is located at the bottom right of image, and the perturbation size is 50/255. The source class and the target class are class 1 and class 3, respectively. Due to that the perturbation size is small and there is only one pixel being perturbed, 900 backdoor training images are required to poison the training set to ensure an effective attack.

For the attack on GTSRB, we use the same ``local square'' pattern used in the main experiments on CIFAR-10. The source classes and the target class are class $\{13, 14, 15, 16, 17,\\ 20, 24, 25, 27, 28, 29, 30\}$ and class 8, respectively. Class 8 images are for 120kph speed limit sign and the picked source classes contain images for traffic signs requiring a slowing down or stopping. In practice, such an attack may cause an autonomous vehicle to speed up while seeing, e.g., a stop sign. For this attack, we create 30 backdoor training images per source class to poison the training set.

For the attack on CIFAR-100, we use the same ``chessboard'' pattern as in \cite{Post-ICASSP}. The source classes and the target class are class $\{1, \cdots, 15\}$ and class 21, respectively. 50 backdoor training images per source class are used to poison the training set.

\begin{figure}[t]
	\centering
	\begin{minipage}[b]{\linewidth}
		\centering
		\includegraphics[width=0.32\linewidth]{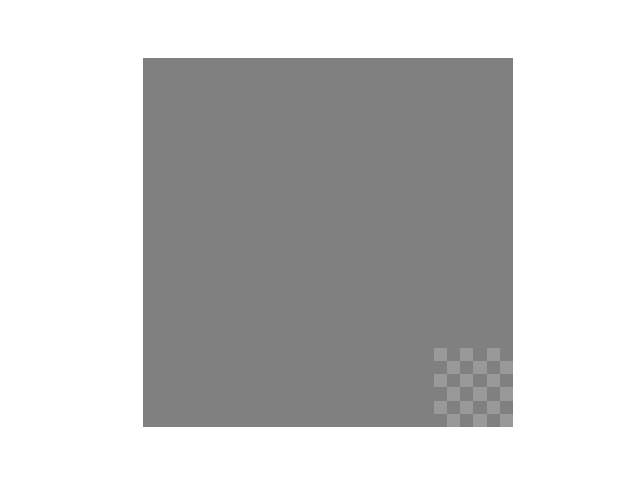}
		\includegraphics[width=0.32\linewidth]{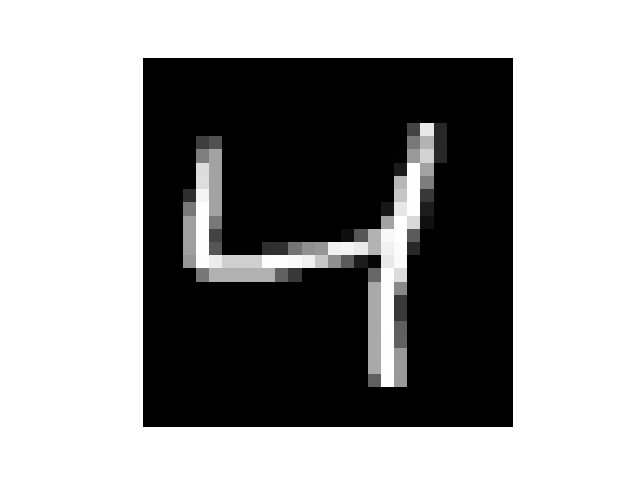}
		\includegraphics[width=0.32\linewidth]{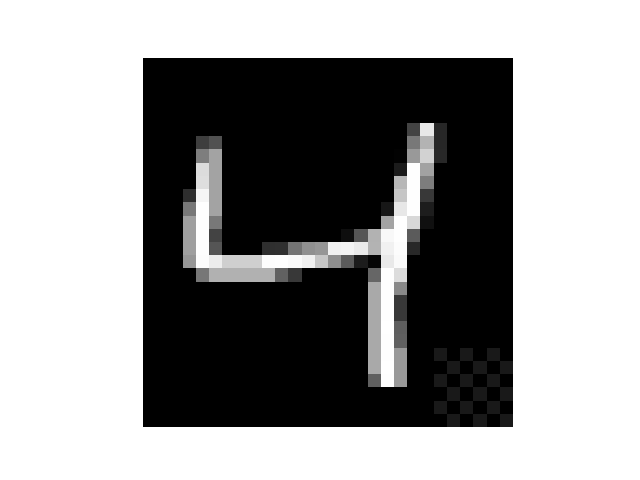}
		\subcaption{MNIST}
	\end{minipage}
	\begin{minipage}[b]{\linewidth}
		\centering
		\includegraphics[width=0.32\linewidth]{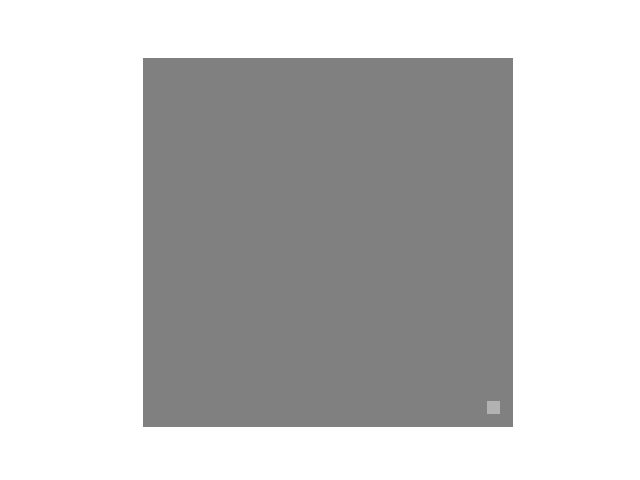}
		\includegraphics[width=0.32\linewidth]{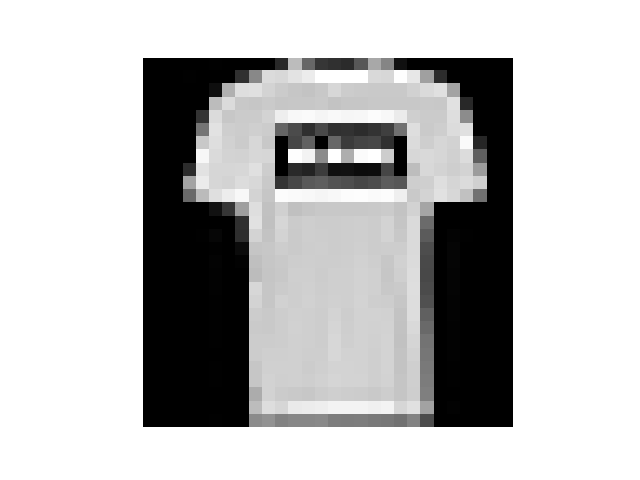}
		\includegraphics[width=0.32\linewidth]{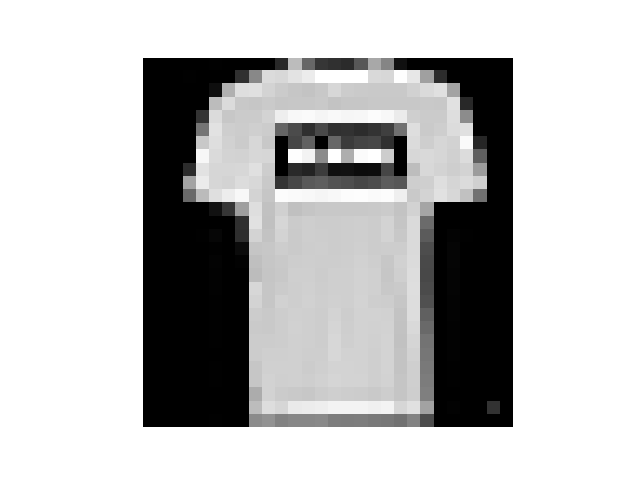}
		\subcaption{F-MNIST}
	\end{minipage}
	\begin{minipage}[b]{\linewidth}
		\centering
		\includegraphics[width=0.32\linewidth]{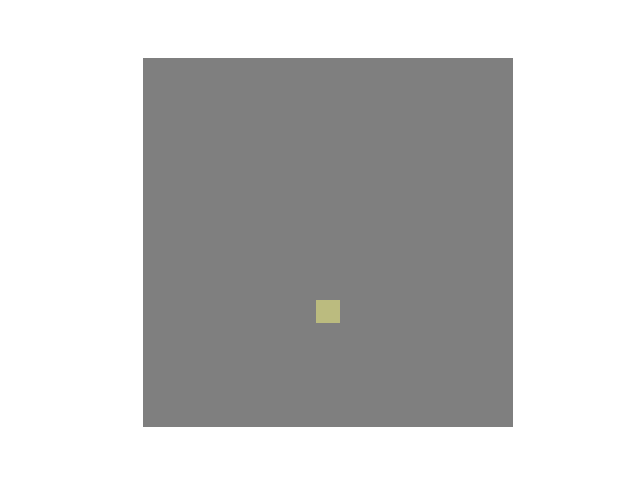}
		\includegraphics[width=0.32\linewidth]{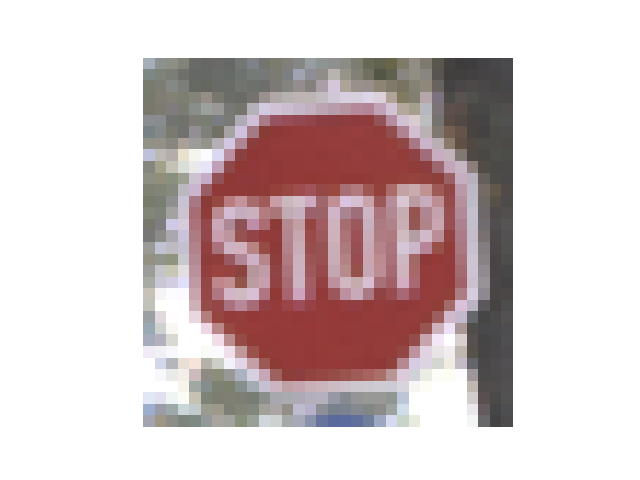}
		\includegraphics[width=0.32\linewidth]{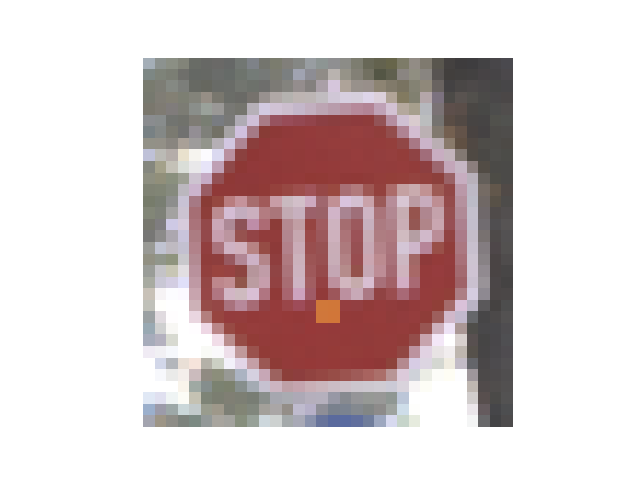}
		\subcaption{GTSRB}
	\end{minipage}
	\begin{minipage}[b]{\linewidth}
		\centering
		\includegraphics[width=0.32\linewidth]{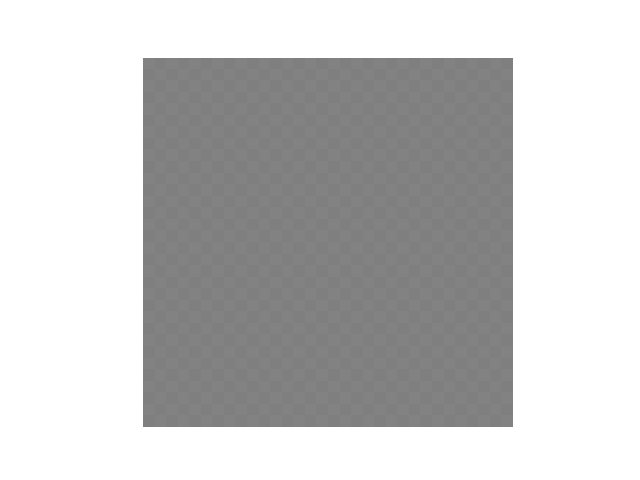}
		\includegraphics[width=0.32\linewidth]{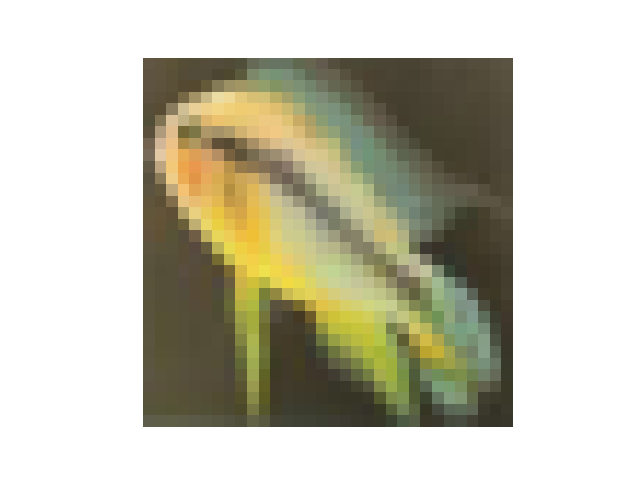}
		\includegraphics[width=0.32\linewidth]{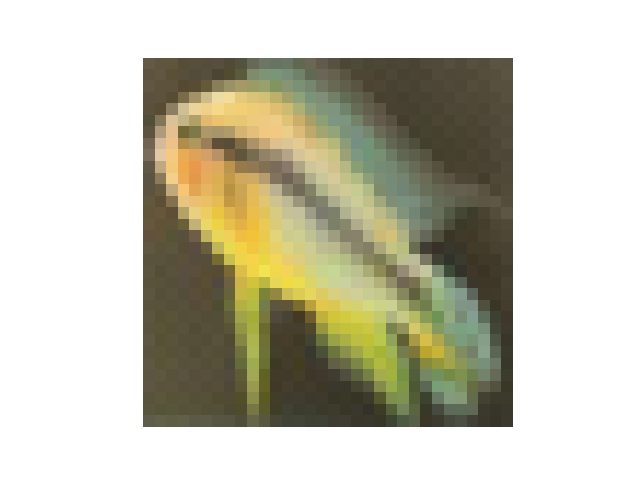}
		\subcaption{CIFAR-100}
	\end{minipage}
	\caption{Backdoor pattern (with 0.5 offset) (left), example backdoor training image (right), and originally clean image (middle), for attacks on MNIST, F-MNIST, GTSRB, and CIFAR-100.}
	\label{fig:bd_example_other_datasets}
\end{figure}

Example backdoor patterns, backdoor training images, and their original clean images for the four attacks are shown in Fig. \ref{fig:bd_example_other_datasets}.

\subsubsection{Training}

We use LeNet-5 \cite{MNIST} DNN architecture for training both the attacked and the clean classifier on MNIST. Training is performed for 60 epochs, with batch size 256 and learning rate $10^{-2}$, without data augmentation. For F-MNIST, we use a ``VGG-9'' DNN architecture, which is customized by removing the last two convolutional layers of VGG-11 \cite{VGG} and using 16, 32, 64, 64, 128, 128 filters in the six remaining convolutional layers, respectively. Training is performed for 60 epochs with batch size 256 and learning rate $10^{-2}$, without data augmentation. For GTSRB, we use the same DNN architecture used in \cite{NC}. Training is performed for 100 epochs with batch size 64 and learning rate $10^{-3}$, without data augmentation. For CIFAR-100, we use the standard ResNet-34 architecture \cite{ResNet}. Training is performed for 120 epochs with batch size 32 and learning rate $10^{-3}$. The same training data augmentation options for training in the main experiments are used here. In Tab. \ref{tab:ASR_CTA_other_datasets}, we show the ASR and CTA for each classifier being attacked, and CTA for the clean classifiers. Clearly, all the attacks are successful with the high ASR and almost no degradation in CTA compared to the clean benchmarks.

\begin{figure}[t]
	\centering
	\begin{minipage}[b]{0.5\linewidth}
		\centering
		\includegraphics[width=\linewidth]{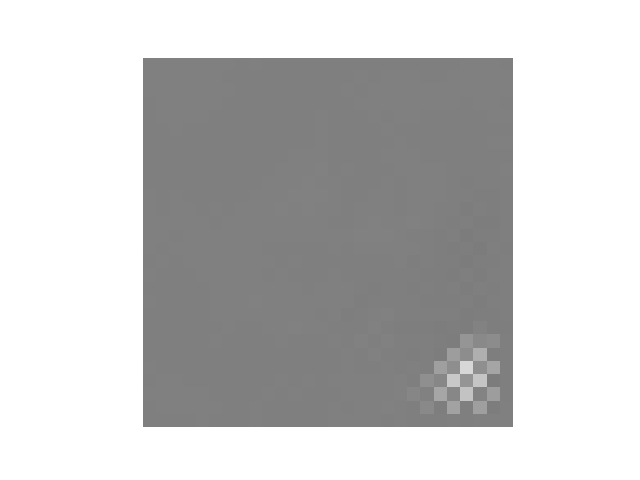}
		\subcaption{MNIST}
	\end{minipage}%
	\begin{minipage}[b]{0.5\linewidth}
		\centering
		\includegraphics[width=\linewidth]{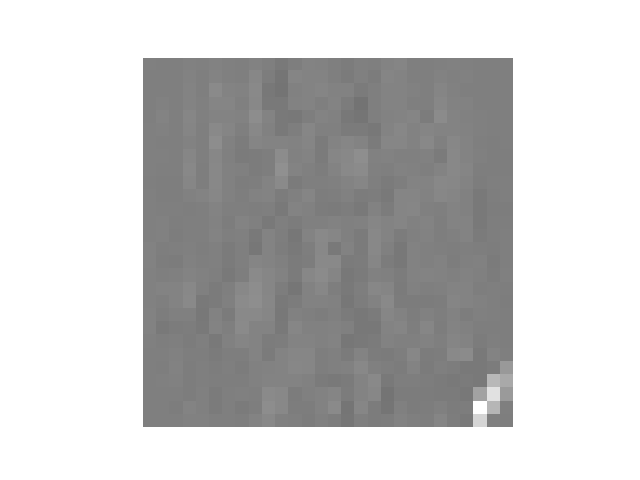}
		\subcaption{F-MNIST}
	\end{minipage}
	\begin{minipage}[b]{0.5\linewidth}
		\centering
		\includegraphics[width=\linewidth]{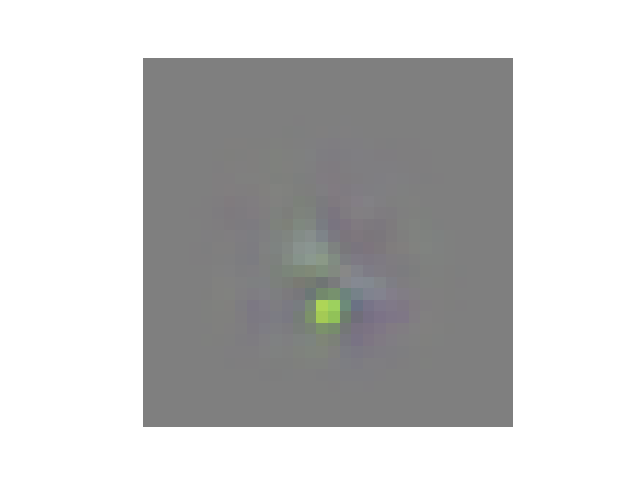}
		\subcaption{GTSRB}
	\end{minipage}%
	\begin{minipage}[b]{0.5\linewidth}
		\centering
		\includegraphics[width=\linewidth]{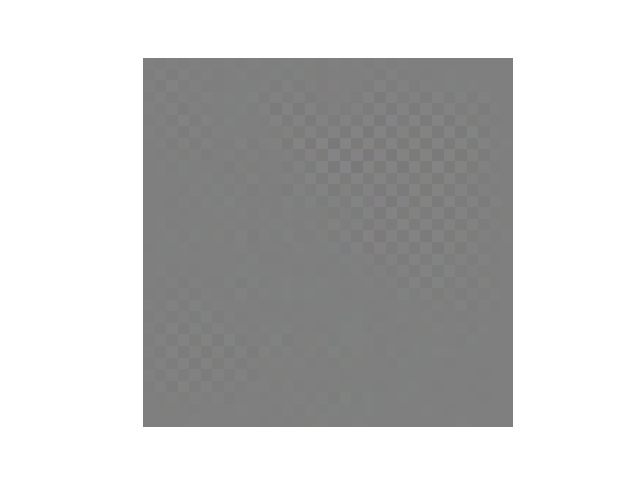}
		\subcaption{CIFAR-100}
	\end{minipage}
	\caption{Backdoor patterns estimated by our defense for attacks on MNIST, FMNIST, GTSRB, and CIFAR-100.}
	\label{fig:bd_est_other_datasets}
\end{figure}

\begin{table}[t]
	\begin{center}
		\caption{Attack success rate (ASR) and clean test accuracy (CTA) of the classifiers trained on the backdoor poisoned training set; and CTA of the clean classifiers for MNIST, F-MNIST, GTSRB, and CIFAR-100.}
		\resizebox{0.45\textwidth}{!}{
			\begin{tabular}{ |c|c|c|c|c| }
				\hline 
				& MNIST & F-MNIST & GTSRB & CIFAR-100\\
				\hline
				attack ASR & 96.3 & 91.7 & 87.8 & 98.0\\
				\hline
				attack CTA & 98.9 & 91.0 & 98.1 & 72.4\\
				\hline
				clean CTA & 99.1 & 90.8 & 98.7 & 72.8\\
				\hline
			\end{tabular}\label{tab:ASR_CTA_other_datasets}}
	\end{center}
\end{table}

\subsubsection{Defense}

As shown in Tab. \ref{tab:pvalues} of the main paper, our defense successfully detects all the four attacks with no false detection when there is no attack. Here, we show the estimated pattern for each detected attack in Fig. \ref{fig:bd_est_other_datasets}. The main features of the backdoor patterns being used are recovered for all the four attacks. Also, the target class for each attack is correctly inferred by our defense.

\end{document}